\documentclass{article}

\usepackage{arxiv}
\usepackage{hyperref}
\usepackage{todonotes}

\usepackage[utf8]{inputenc} 
\usepackage[T1]{fontenc}    
\usepackage{hyperref}       
\usepackage{url}            
\usepackage{booktabs}       
\usepackage{amsfonts}       
\usepackage{nicefrac}       
\usepackage{microtype}      
\usepackage{lipsum}
\usepackage{footnote}
\usepackage{booktabs}
\makesavenoteenv{table}

\title{WikiDataSets : \\Standardized sub-graphs from Wikidata}

\author{
  Armand Boschin\\
  Télécom Paris\\
  Institut Polytechnique de Paris\\
  Paris, France\\
  \texttt{aboschin@enst.fr} \\
  \And
  Thomas Bonald \\
  Télécom Paris\\
  Institut Polytechnique de Paris\\
  Paris, France\\
  \texttt{tbonald@enst.fr} \\
}

\begin{document}
\maketitle

\begin{abstract}
Developing new ideas and algorithms in the fields of graph processing and relational learning requires public datasets. While Wikidata is the largest open source knowledge graph, involving more than fifty million entities, it is larger than needed in many cases and even too large to be processed easily. Still, it is a goldmine of relevant facts and relations. Using this knowledge graph is time consuming and prone to task specific tuning which can affect reproducibility of results. Providing a unified framework to extract topic-specific subgraphs solves this problem and allows researchers to evaluate algorithms on common datasets. This paper presents various topic-specific subgraphs of Wikidata along with the generic Python code used to extract them. These datasets can help develop new methods of knowledge graph processing and relational learning.
\end{abstract}


\section{Motivation}

Relational learning as been a hot topic for a couple of years. The most widespread datasets used for benchmarking new methods (\cite{bordes_translating_2013} \cite{wang_knowledge_2014} \cite{ji_knowledge_2015}, \cite{trouillon_complex_2017}, \cite{dettmers_convolutional_2017}, \cite{nguyen_novel_2018}) are a subset of Wordnet (WN18) and two subsets of Freebase (FB15k, FB1M). Recently, Wikidata has been used in the literature, as a whole in \cite{lerer_pytorch-biggraph:_2019} or as subsets (about people and films) in \cite{ostapuk_activelink:_2019,guan_link_2019}. It is a diverse and qualitative dataset that could useful for many researchers. In order to make it easier to use, we decided to build thematic subsets which are publicly available\footnote{\url{https://graphs.telecom-paristech.fr/}}. They are presented in this paper.

\section{Overview of the datasets}

A knowledge graph is given by a set of nodes (entities) and a set of facts linking those nodes. A fact is a triplet of the form $(head, relation, tail)$ linking two nodes by a typed edge (\textit{relation}). Wikidata is a large knowledge graph, archives of which can be downloaded.

We propose five topic-related datasets which are subgraphs of the Wikidata knowledge-graph : \texttt{animal} \texttt{species}, \texttt{companies}, \texttt{countries}, \texttt{films} and \texttt{humans}. Each graph includes only nodes that are instances of its topic. For example, the dataset \texttt{humans} contains the node \textit{George Washington} because the fact (\textit{George Washington}, \textit{isInstanceOf}, \textit{human}) is true in Wikidata.

More exactly, to be included nodes should be instances of the topic or of any sub-class of this topic. For example, \texttt{countries} contains \textit{USSR} though \textit{USSR} is not an instance of \textit{country}. It is however an instance of \textit{historical country} which is a subclass of \textit{country}. Topics and example of the corresponding sub-classes are reported in Table \ref{topics}.

Each graph includes as edges only the relations that stand true in Wikidata. For example, the dataset \texttt{humans} contains the nodes \textit{George Whashington} and \textit{Martha Washington} and there is an edge between the two as the fact (\textit{George Whashington}, \textit{spouse}, \textit{Martha Washington}) is true.

Eventually, Wikidata facts linking the selected nodes to Wikidata entities that were not selected (because not instances of the topic)  are also kept as attributes of the nodes. Note that as all the edges are labeled, each graph is a knowledge-graph itself.

For each dataset, some metadata is provided in Table \ref{metadata} along with a couple of examples of nodes, edges and attributes in Table \ref{examples}, the distributions of the edge types in Tables \ref{edges_animals}, \ref{edges_companies}, \ref{edges_countries}, \ref{edges_films}, \ref{edges_humans} and details of the files in Table \ref{ffiles}.

\section{Presentation of the code}
The proposed datasets were built using the \href{https://pypi.org/project/wikidatasets/}{WikiDataSets} package, which was made openly available on PyPI\footnote{\url{https://pypi.org/project/wikidatasets/}}. It is also documented online. The three following steps are necessary.

\begin{enumerate}
    \item Call to  \texttt{get\_subclasses} on the Wikidata ID of the topic entity (e.g. Q5 for humans) to fetch the various Wikidata entities which are sub-classes of the topic entity (e.g. sub-classes of humans).
    \item Call to \texttt{query\_wikidata\_dump} to read each line of the Wikidata archive dump \texttt{latest-all.json.bz2}\footnote{\url{https://dumps.wikimedia.org/wikidatawiki/entities/}} and keep only the lines corresponding to selected nodes. This returns a list of facts stored as pickle files. Labels of the entities and relations are also collected on the way to be able to provide the labels of the attributes. Those labels are collected in English when available.
    \item Eventually \texttt{build\_dataset} turns this list of facts into the five files presented in table \ref{ffiles} (facts between nodes, attributes, entity dictionary giving for each entity its label and its Wikidata ID, relation dictionary giving for each relation its label and its Wikidata id).
\end{enumerate}

\section{Use case examples}

\subsection{Community Detection}
Communities were detected in the humans dataset with the Louvain algorithm \cite{blondel_fast_2008}.

\begin{figure}
    \centering
    \includegraphics[width=10cm]{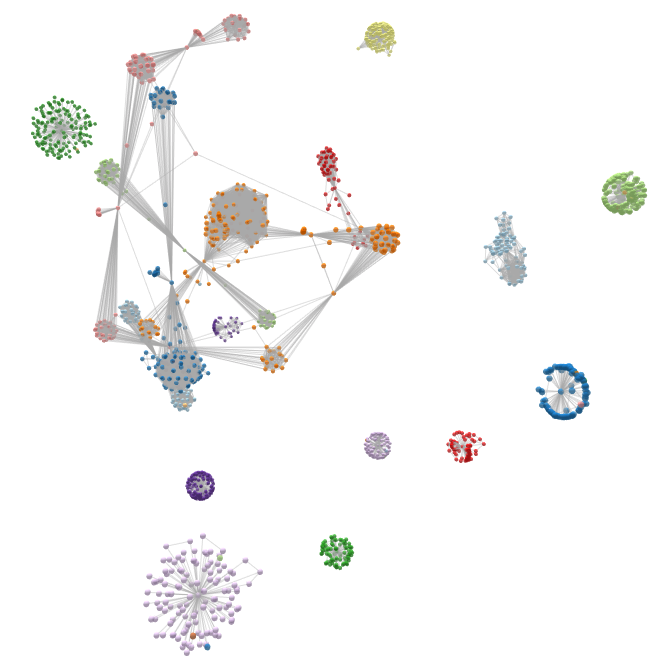}
    \caption[Caption for LOF]{Results of the Louvain algorithm on \texttt{humans} dataset. Each node has been assigned a community and each color corresponds to a community.}
    \label{fig:comm}
\end{figure}

Using the Scikit-Network\footnote{\url{https://pypi.org/project/scikit-network/}} framework, we extracted communities from the \texttt{humans} dataset with the Louvain algorithm \cite{blondel_fast_2008}. In order to visualize the communities, the 50 nodes of highest degree were then extracted along with their neighbors and were represented using the 3d-force-graph library\footnote{\url{https://github.com/vasturiano/3d-force-graph}}. A snapshot of the visualization is presented in Figure~\ref{fig:comm}.

Navigating through the graph, we find communities that seem to make sense (e.g. American artists, Vietnamese political leaders). We find as well (in pink on Figure \ref{fig:comm}) the Chinese Tang dynasty, each small \textit{ball} corresponding to an emperor and its wife and children.

\subsection{Knowledge Graph Embedding}
As noted before, \texttt{humans} is a knowledge graph and it can be embedded using off-the-shelf methods. TransH and ANALOGY were tested using the same hyper-parameters as the ones recommended for the FB15k dataset in the original papers \cite{wang_knowledge_2014,liu_analogical_2017}.

Attributes are not included in the process and nodes with less than 5 neighbors are filtered out. The facts are then randomly split into training (0.8) and testing (0.2) sets. Training was done using the TorchKGE\footnote{\url{https://pypi.org/project/torchkge/}} library on Nvidia Titan~V GPU during 1,000 epochs. The embedding quality was evaluated on a link prediction task as in \cite{bordes_translating_2013}and results are presented in Table~\ref{results}. Let us note that these scores are quite high but this is mainly because entities involved in less than five facts were filtered out.

\begin{table}
    \centering
    \caption{Topics and examples of their sub-classes for each dataset.}
    \label{topics}
    \begin{tabular}{|l||l|l|l|}
\hline

Dataset   & topic & Wikidata ID & sub-class examples \\ \hline \hline
\texttt{animals} & \texttt{taxon} & \href{https://www.wikidata.org/wiki/Q16521}{Q16521} & reptilia classifications, amphibia classifications\\ \hline
\texttt{companies} & \texttt{business} & \href{https://www.wikidata.org/wiki/Q4830453}{Q4830453} &  low-cost airline, fast food chain\\ \hline
\texttt{countries} & \texttt{country} & \href{https://www.wikidata.org/wiki/Q6256}{Q6256} &  arab caliphate, colonial empire \\ \hline
\texttt{films} & \texttt{film} & \href{https://www.wikidata.org/wiki/Q11424}{Q11424} & cartoon, apocalyptic movie \\ \hline
\texttt{humans} & \texttt{human} & \href{https://www.wikidata.org/wiki/Q5}{Q5} & human being, child, patient \\ \hline
\end{tabular}
\end{table}

\begin{table}
    \centering
    \caption{Metadata of each dataset.}
    \label{metadata}
    \begin{tabular}{|l||l|l|l|l|l|l|l|}
\hline

Dataset   & \# nodes & \# edges & \begin{tabular}[c]{@{}c@{}}\# isolated \\ nodes\footnote{Nodes that are isolated from the rest of the graph but have attributes.}\end{tabular} & \begin{tabular}[c]{@{}c@{}}\# distinct \\ attributes\end{tabular}& \begin{tabular}[c]{@{}c@{}}\# attribute \\ facts\end{tabular} & \begin{tabular}[c]{@{}c@{}}\# distinct \\ relations\end{tabular}& \begin{tabular}[c]{@{}c@{}}\# distinct attribute \\ relations\end{tabular} \\ \hline \hline
\texttt{animals} & 2,617,023 & 2,747,853 & 19,574 & 164,477 & 563,7811 & 45 & 122 \\ \hline
\texttt{companies} & 249,619 & 50,475 & 216,421 & 94,526 & 814,068 & 101 & 339 \\ \hline
\texttt{countries} & 3,324 & 10,198 & 1,778 & 17,965 & 37,096 & 48 & 145 \\ \hline
\texttt{films} & 281,988 & 9,221 & 274,062 & 236,187 & 3,104,681 & 44 & 236 \\ \hline
\texttt{humans} & 5,043,535 & 1,059,144 & 4,538,119 & 774,494 & 34,848,421 & 165 & 439 \\ \hline
\end{tabular}
\end{table}

\begin{table}
\centering
\caption{Examples of nodes, facts and attributes for each dataset.}
\label{examples}
\begin{tabular}{|l||c|c|c|}
\hline
dataset & \multicolumn{1}{l|}{Example of node} & \multicolumn{1}{l|}{Example of edge} & \multicolumn{1}{l|}{Example of attribute} \\ \hline \hline
\texttt{animals} & Tiger & \begin{tabular}[c]{@{}c@{}}(Tiger, \\ parent taxon, \\ Panthera)\end{tabular} & \begin{tabular}[c]{@{}c@{}}(Tiger, \\ produced sound,\\ roar)\end{tabular} \\ \hline
\texttt{companies} & Deutsche TeleKom & \begin{tabular}[c]{@{}c@{}}(Deutsche TeleKom, \\ subsidiary, \\ T-Mobile US)\end{tabular} & \begin{tabular}[c]{@{}c@{}}(Deutsche Telecom, \\ award received,\\ Big Brother Award)\end{tabular} \\ \hline
\texttt{countries} & France & \begin{tabular}[c]{@{}c@{}}(France,\\ shares border with, \\ Switzerland)\end{tabular}    & \begin{tabular}[c]{@{}c@{}}(France,\\currency,\\ euro)\end{tabular} \\ \hline
\texttt{films} & Fast Five & \begin{tabular}[c]{@{}c@{}}(Fast Five,\\ Follows, \\ Fast \& Furious)\end{tabular} & \begin{tabular}[c]{@{}c@{}}(Fast Five,\\cast member,\\Paul Walker)\end{tabular} \\ \hline
\texttt{humans} & George Washington & \begin{tabular}[c]{@{}c@{}}(George Washington,\\spouse,\\Martha Washington)\end{tabular} & \begin{tabular}[c]{@{}c@{}}(George Washington,\\cause of death,\\ Epiglottitis)\end{tabular}\\ \hline
\end{tabular}
\end{table}

\begin{table}
    \centering
    \caption{Details of the files for each dataset.}
    \label{ffiles}
    \begin{tabular}{|l|l|}
\hline 
File           & Description                                                                       \\ \hline \hline
\texttt{readme.txt}     & Contains meta-data about the dataset                                              \\ \hline
\texttt{attributes.txt} & List facts linking nodes to their attributes (in the form from, to, rel). \\ \hline
\texttt{edges.txt}      & List of facts linking nodes between them (in the form from, to, rel).     \\ \hline
\texttt{entities.txt}  & Dictionary linking entities to their Wikidata codes and labels.                   \\ \hline
\texttt{nodes.txt}  & Subset of \texttt{entities.txt} containing only nodes of the graph                  \\ \hline
\texttt{relations.txt}  & Dictionary linking relations to their Wikidata codes and labels.                  \\ \hline
\end{tabular}
\end{table}

\begin{table}
\centering
    \caption{Distribution of the top 20 edge types in the \texttt{animals} dataset.}
    \label{edges_animals}
\begin{tabular}{lr}
\toprule
label &  headEntity \\
\midrule
parent taxon                                &     2604202 \\
basionym                                    &       97381 \\
taxonomic type                              &       22949 \\
original combination                        &        9077 \\
taxon synonym                               &        6967 \\
different from                              &        2578 \\
this zoological name is coordinate with     &        1885 \\
host                                        &         994 \\
parent of this hybrid, breed, or cultivar   &         470 \\
replaced synonym (for nom. nov.)            &         367 \\
subclass of                                 &         179 \\
said to be the same as                      &         162 \\
afflicts                                    &         129 \\
instance of                                 &          88 \\
based on                                    &          83 \\
named after                                 &          53 \\
has part                                    &          53 \\
derivative work                             &          42 \\
main food source                            &          33 \\
follows                                     &          17 \\
followed by                                 &          17 \\
\bottomrule
\end{tabular}

\end{table}

\begin{table}
\centering
    \caption{Distribution of the top 20 edge types in the \texttt{companies} dataset.}
    \label{edges_companies}
\begin{tabular}{lr}
\toprule
label &  headEntity \\
\midrule
parent organization     &       11850 \\
owned by                &       10783 \\
subsidiary              &        8574 \\
owner of                &        4669 \\
instance of             &        2840 \\
followed by             &        1726 \\
follows                 &        1637 \\
replaced by             &        1300 \\
replaces                &        1053 \\
operator                &         993 \\
business division       &         569 \\
member of               &         565 \\
different from          &         557 \\
founded by              &         536 \\
stock exchange          &         528 \\
part of                 &         444 \\
architect               &         237 \\
has part                &         190 \\
manufacturer            &         146 \\
distributor             &         144 \\
industry                &         114 \\
\bottomrule
\end{tabular}
\end{table}

\begin{table}
\centering
    \caption{Distribution of the top 20 edge types in the \texttt{countries} dataset.}
    \label{edges_countries}
\begin{tabular}{lr}
\toprule
label &  headEntity \\
\midrule
diplomatic relation                                 &        6009 \\
shares border with                                  &        1202 \\
country                                             &        1090 \\
replaced by                                         &         362 \\
replaces                                            &         323 \\
follows                                             &         263 \\
followed by                                         &         263 \\
located in the administrative territorial entity    &         157 \\
has part                                            &         106 \\
part of                                             &          89 \\
different from                                      &          74 \\
contains administrative territorial entity          &          60 \\
capital                                             &          24 \\
capital of                                          &          19 \\
territory claimed by                                &          18 \\
separated from                                      &          14 \\
instance of                                         &          11 \\
facet of                                            &          11 \\
named after                                         &          10 \\
headquarters location                               &          10 \\
founded by                                          &           9 \\
\bottomrule
\end{tabular}
\end{table}

\begin{table}
\centering
    \caption{Distribution of the top 20 edge types in the \texttt{films} dataset.}
    \label{edges_films}
\begin{tabular}{lr}
\toprule
label                           &  headEntity \\
\midrule
follows                         &        3070 \\
followed by                     &        3049 \\
different from                  &         910 \\
based on                        &         703 \\
has part                        &         399 \\
part of                         &         348 \\
derivative work                 &         271 \\
said to be the same as          &         138 \\
part of the series              &         108 \\
cast member                     &          59 \\
inspired by                     &          29 \\
main subject                    &          17 \\
producer                        &          12 \\
has edition                     &          10 \\
edition or translation of       &          10 \\
genre                           &           8 \\
production company              &           7 \\
influenced by                   &           7 \\
award received                  &           6 \\
distributor                     &           5 \\
named after                     &           5 \\
\bottomrule
\end{tabular}
\end{table}

\begin{table}
\centering
    \caption{Distribution of the top 20 edge types in the \texttt{humans} dataset.}
    \label{edges_humans}
\begin{tabular}{lr}
\toprule
label &  headEntity \\
\midrule
child                           &      255885 \\
sibling                         &      241225 \\
father                          &      212661 \\
spouse                          &      123117 \\
mother                          &       44035 \\
student of                      &       29931 \\
different from                  &       27395 \\
student                         &       25578 \\
doctoral advisor                &       20315 \\
relative                        &       19844 \\
doctoral student                &       17630 \\
consecrator                     &       11119 \\
influenced by                   &        6936 \\
professional or sports partner  &        6230 \\
partner                         &        6197 \\
head coach                      &        4248 \\
employer                        &        1309 \\
said to be the same as          &         670 \\
killed by                       &         477 \\
sponsor                         &         474 \\
replaces                        &         347 \\
\bottomrule
\end{tabular}
\end{table}

{}

\begin{table}
\centering
\caption{Performances of TransH and ANALOGY models on \texttt{humans} dataset filtered to keep only
entities involved in more than 5 facts resulting in 238,376 entities and 722,993 facts.}
\label{results}
\begin{tabular}{|l||l|l|l|l|}
\hline
Model   & Hit@10    & Filt. Hit@10  & Mean Rank & Filt. Mean Rank \\[2pt] \hline \hline
TransH  & 51.5      & 61.8          & 15,030    & 15,027 \\ \hline
ANALOGY & 59.6      & 67.9          & 10,149    & 10,147 \\ \hline
\end{tabular}
\end{table}

\bibliographystyle{unsrt}  
\bibliography{references}
\end{document}